\documentclass[a4paper]{svproc}

\usepackage{url}

\usepackage{cite}
\usepackage{amsmath,amssymb,amsfonts}
\usepackage{algorithmic}
\usepackage{graphicx}
\usepackage{textcomp}
\usepackage{xcolor}
\usepackage{siunitx}
\usepackage{float}
\usepackage[symbol]{footmisc}
\usepackage{ar}
\usepackage{balance}
\usepackage[hidelinks]{hyperref}
\usepackage{url}
\hypersetup{
    colorlinks=true,
    linkcolor=black,
    filecolor=none,      
    urlcolor=blue,
    citecolor=black,
    pdftitle={Overleaf Example},
    pdfpagemode=FullScreen,
    }
\renewcommand{\thefootnote}{\fnsymbol{footnote}}

\newcommand\blfootnote[1]{%
  \begingroup
  \renewcommand\thefootnote{}\footnote{#1}%
  \addtocounter{footnote}{-1}%
  \endgroup
}

\newcommand{\bs}[1]{\boldsymbol{#1}}
\renewcommand{\ts}[1]{\text{#1}}

\title{\mbox{Power-Efficient} Actuation for \mbox{Insect-Scale} Autonomous Underwater Vehicles}

\titlerunning{Power-Efficient Actuation for \mbox{Insect-Scale} AUVs}  

\author{Cody R. Longwell \and Conor K. Trygstad \and N\'estor O. P\'erez-Arancibia}
\authorrunning{C. R. Longwell et al.} 
\institute{Washington State University, Pullman, WA\,99164-2920, USA\\
\email{cody.longwell@wsu.edu},
\email{conor.trygstad@wsu.edu}, \email{n.perezarancibia@wsu.edu}}

\begin{document}
\mainmatter             

\sloppy

\maketitle          

\begin{abstract}
We present a new evolution of the \textit{Very Little \mbox{Eel-Inspired} roBot}, \mbox{the~VLEIBot\textsuperscript{++}}\hspace{-0.4ex}, a \mbox{$900$-mg} swimmer propelled by two \mbox{$10$-mg} bare \textit{high-work-density} (HWD) actuators driven by \textit{shape-memory \mbox{alloy}} (SMA) wires. An actuator of this type consumes an average power of about $40\,\ts{mW}$ during \mbox{in-air} operation. We integrated onboard power and computation by using a \mbox{custom-built} \textit{printed circuit board} (PCB) and an \mbox{11-mAh} \mbox{$3.7$-V} \mbox{$507$-mg} \mbox{single-cell} \textit{\mbox{lithium-ion}} (\mbox{Li-Ion}) battery, which in conjunction enable autonomous swimming for about \mbox{$20\,\ts{min}$} on a single charge. The \mbox{VLEIBot\textsuperscript{++}\hspace{-0.3ex}} can swim at speeds of up to \mbox{$18.7\,\ts{mm}/\ts{s}$}~\mbox{($0.46\,\ts{Bl}/\ts{s}$)} and is the first subgram microswimmer with onboard power, actuation, and computation developed to date. Unfortunately, the approach employed to actuate this robot is infeasible for underwater applications because a typical \mbox{$10$-mg} bare \mbox{SMA-based} actuator requires an average power of approximately \mbox{$800\,\ts{mW}$} when operating underwater. To address this issue, we introduce a new \mbox{$13$-mg} \mbox{power-efficient} \mbox{SMA-based} actuator that can function with similar power requirements (\mbox{$\sim\hspace{-0.4ex}80\,\ts{mW}$} on average) and output performance (\mbox{$\sim\hspace{-0.4ex}3\,\ts{mm}$} at low frequencies) in air and water. This design uses a sealed flexible \mbox{air-capsule} to enclose the SMA wires that drive the actuator and thus passively control the \mbox{heat-transfer} rate of the thermal system. Furthermore, this new \mbox{power-efficient} encapsulated actuator requires low voltages of excitation (\mbox{$3$~to~$4\,\ts{V}$}) and simple power electronics to function. The breakthroughs presented in this paper represent a path towards the creation of \mbox{insect-scale} \textit{autonomous underwater vehicles} (AUVs).\blfootnote{C.\,R.\,Longwell and C.\,K.\,Trygstad contributed equally to this research.}
\end{abstract}

\section{An Aquatic Frontier for Microrobotics} 
\label{SECTION01}
\vspace{-2ex}
The recent introduction of new microfabrication methods, actuation materials, propulsion mechanisms, and control approaches for microrobotics has enabled the development of several novel \mbox{high-performance} aquatic~\cite{ Waterstrider_2023,VLEIBot_2024,FRISHBot_2024,Li2023PZT}, aerial~\cite{BeePlus_2019, BeePlusPlus_2023, ren2022DEAFlyer}, and terrestrial~\cite{wu2019insect, gravish2020stcrawler, SMALLBug_2020, SMARTI_2021} \mbox{insect-scale} robots. We envision a future in which swarms of triphibious microrobots work in coordination with humans to assist in search and rescue operations, pollution detection and mitigation, infrastructure and facility inspections, and automated pest control, just to mention a few possibilities. For this vision to become a reality, the deployed microrobots must operate autonomously in unstructured fields and adapt to environmental uncertainties. Currently, the majority of microrobotic platforms either function physically tethered to power sources~\cite{BeePlus_2019, BeePlusPlus_2023, wu2019insect, gravish2020stcrawler, SMALLBug_2020, SMARTI_2021, Waterstrider_2023, FRISHBot_2024, VLEIBot_2024,ren2022DEAFlyer, Li2023PZT, Chen2017HybridBioinspiredVehicle} or rely on large and complex external actuation apparatuses that are impractical for use in \mbox{real-life} situations~\cite{Xu2013HelicalMicroswimmers, Zhang2010RotatingNickleNanowires, Liu2010WirelessSwimmingMicrorobot}. While the creation of novel actuation and \mbox{energy-storage} technologies has contributed to the realization of new \mbox{mm-to-cm--scale} autonomous terrestrial vehicles~\cite{goldberg2018power, MilliMobile2023, Ji2019DEAnsect, RoBeetle_2020}, these technologies have not yet been leveraged to develop \mbox{insect-sized} \textit{\mbox{autonomous} underwater vehicles} (AUVs). On the other hand, a significant number of \mbox{cm-to-dm--scale} underwater vehicles showing great promise for deployment in unstructured fields have been produced in the past two decades~\cite{Spino2024Bomb,Berlinger2021BlueSwarm,Berlinger2018DEAFish,Berlinger2021AutonomousFish,Li2024MiniSub,Cen2013Piezohydroelastic,Wang2023Jellyfish,Giorgio2013SquidHead,Villanueva2009SMAJellyfish}; however, all these platforms use actuation methods and power sources that are not easily scalable to \mbox{mm-to-cm} sizes. 
\begin{figure}[t!]
\vspace{-0.2ex}
\begin{center}    
\includegraphics[width=0.98\textwidth]{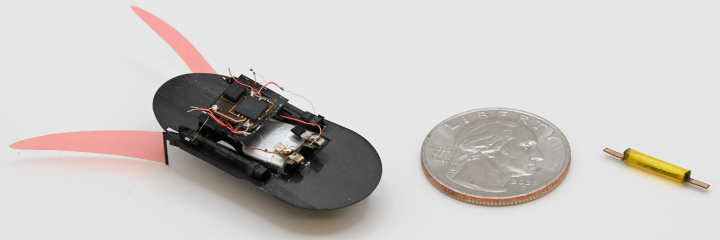}
\end{center}
\vspace{-4ex}
\caption{\textbf{A swimmer and an actuator.} The \mbox{VLEIBot\textsuperscript{++}\,(\textit{left})}, a \mbox{$900$-mg} \mbox{autonomous} surface swimmer driven by two \mbox{$10$-mg} bare \mbox{SMA-based} \mbox{actuators}; and, a new \mbox{low-power} \mbox{$13$-mg} encapsulated HWD \mbox{SMA-based} actuator for underwater operation (\textit{right}). This new actuator has a length of $15.25\,\ts{mm}$, a volume without the capsule of $2.37\,\ts{mm}^3$, and a volume with the capsule of $33.02\,\ts{mm}^3$. The actuators that drive the \mbox{VLEIBot\textsuperscript{++}} have a length of $12\,\ts{mm}$ and volume of $1.89\,\ts{mm}^3$ each. 
\label{FIG01}}
\vspace{-2ex}
\end{figure}

One approach that has been successfully demonstrated to be \mbox{mm-scalable} is microactuation based on the use of tensioned thin \textit{shape-memory alloy} (SMA) wires~\cite{SMALLBug_2020,SMARTI_2021}. Theoretically, \mbox{SMA-based} microactuation is a promising path towards the development of \mbox{insect-scale} AUVs because it exhibits \textit{high work densities} (HWDs), requires low excitation voltages, is passively controllable using simple structures made of \textit{carbon~fiber} (CF), and can be driven with simple \mbox{battery-excitable} power electronics.
Using this technology, we created the \mbox{VLEIBot\textsuperscript{++}}\hspace{-0.4ex}, the \mbox{$900$-mg} autonomous surface swimmer shown in \mbox{Fig.\,\ref{FIG01}}. This microrobot is an evolution of the swimmers presented in~\cite{VLEIBot_2024} and is driven by \mbox{$10$-mg} \mbox{low-power} bare \mbox{SMA-based} actuators designed to operate in air. During swimming, an onboard \mbox{11-mAh} \mbox{$3.7$-V} \mbox{$507$-mg} \mbox{single-cell} \textit{\mbox{lithium-ion}} (\mbox{Li-Ion}) battery and a \mbox{custom-built} \textit{printed circuit board} (PCB) provide power for onboard computation and actuation. The \mbox{VLEIBot\textsuperscript{++}\hspace{-0.3ex}~is} the first subgram autonomous surface microrobotic swimmer developed to date and can achieve locomotion speeds of up to $18.7\,\ts{mm/s}$ ($0.46\,\ts{Bl/s}$). We believe that these results represent a breakthrough in microrobotics because it is well known that swimming using traditional \mbox{human-scale} methods becomes increasingly difficult as the Reynolds number decreases~\cite{PurcellAJP_1977}. Each bare \mbox{SMA-based} microactuator employed to drive the \mbox{VLEIBot\textsuperscript{++}\hspace{-0.3ex}~consumes} an average power of approximately $40\,\ts{mW}$ only, thus enabling about $20\,\ts{min}$ of operational time on a single battery charge. Unfortunately, this swimmer is currently constrained to function on the surface of water to keep its actuators dry while swimming, as their power efficiency is significantly degraded when excited underwater. Specifically, when operating underwater, these devices require an average power on the order of $800\,\ts{mW}$, which represents an increase of about $1900\,\%$ with respect to the \mbox{in-air} actuation case. These figures indicate that the development of a \mbox{VLEIBot\textsuperscript{++}-like} underwater swimmer electrically driven by bare \mbox{SMA-based} microactuators is infeasible due to the low energy and power densities of even the most advanced batteries currently available. The problem at the core of this issue is that the \mbox{heat-transfer} coefficient of water is significantly larger than that of air at normal conditions of temperature and pressure (\mbox{$20\,^{\circ}\ts{C}$; \mbox{$1\,\ts{atm}$}}). To address this challenge, we introduce a new \mbox{$13$-mg} \mbox{low-power} \mbox{SMA-based} actuator---shown in \mbox{Fig.\,\ref{FIG01}}---which is capable of both \mbox{in-air} and underwater operation with almost identical power requirements; namely, about $80\,\ts{mW}$, on average, during actuation in both media. This measurement represents an improvement of about $90\,\%$ in energy efficiency with respect to that obtained with a bare \mbox{SMA-based} actuator functioning in water. As seen in~\mbox{Fig.\,\ref{FIG01}}, in the proposed design, the SMA wires that drive the actuator are encapsulated using a \mbox{Kapton-made} flexible chamber that passively controls the local rate of heat transfer. This innovation opens a new technological path towards the realization of lightweight \mbox{low-power} actuation for \mbox{insect-scale~AUVs}. 

The rest of the paper is organized as follows. \mbox{Section\,\ref{SECTION02}} describes the design and fabrication of the \mbox{VLEIBot\textsuperscript{++}}\hspace{-0.4ex}, and discusses preliminary untethered swimming experiments. \mbox{Section\,\ref{SECTION03}} presents experimental results regarding the \mbox{power-consumption} characterization of a \mbox{$10$-mg} bare \mbox{SMA-based} actuator, and describes the proposed \mbox{$13$-mg} \mbox{low-power} \mbox{SMA-based} actuator, which is capable of \mbox{power-efficient} underwater operation. Last, \mbox{Section\,\ref{SECTION04}} summarizes the findings produced through the presented research and states future directions for the development of \mbox{insect-scale} AUVs.
\begin{figure}[ht!]
\vspace{1ex}   
\begin{center}
\includegraphics[width=0.98\textwidth]{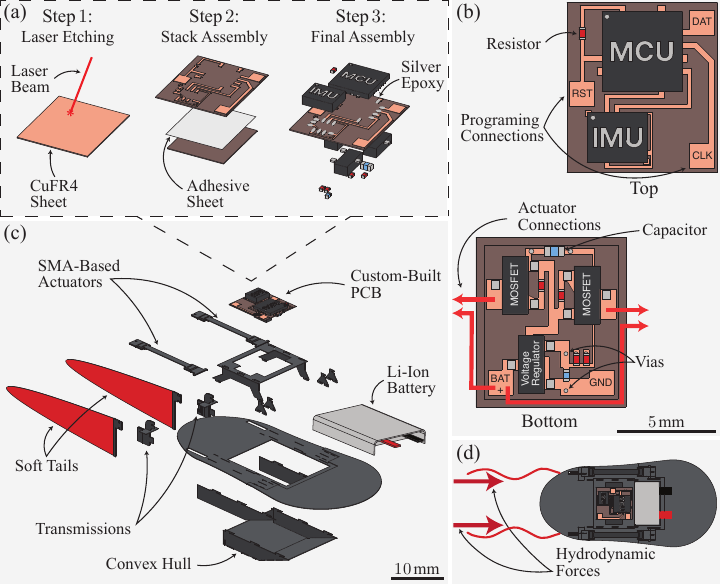}
\end{center}
\vspace{-4ex}
\caption{\textbf{Design and fabrication of the VLEIBot\textsuperscript{++}\hspace{-0.4ex}.} \textbf{(a)}~Fabrication process of the robot's \mbox{custom-designed} PCB. \mbox{In Step\,$1$}, sheets of \mbox{CuFR4} are etched using laser rasterization in order to remove areas of the \mbox{Cu-coating} and thus create the patterns designed for each side of the PCB. \mbox{In Step\,$2$}, the two sides of the PCB are \mbox{pin-aligned} and adhered together with a sheet of Pyralux adhesive by applying pressure and heat inside a curing oven. \mbox{In Step\,$3$}, the \mbox{off-the-shelf} elements composing the interconnected power, actuation, and computation circuits are installed on the two sides of the PCB using conductive silver epoxy (\mbox{MG-Chemicals~$8331$D}) and cured inside an oven. \textbf{(b)}~Top and bottom of the robot's PCB. The top side of the PCB supports an MCU for computation and an IMU for sensing. The bottom side of the PCB supports two \mbox{$\ts{N}$-channel} \mbox{MOSFETs} and a voltage regulator. The two sides of the PCB are connected through \textit{vias}. Decoupling capacitors and \mbox{pull-up/down} resistors are necessary to ensure the functionality of the circuits. \textbf{(c)}~Exploded view of the VLEIBot\textsuperscript{++}'s design. This robot is composed of six main types of components: (i)~two bare \mbox{SMA-based} microactuators of the type described in~\cite{VLEIBot_2024} with passive hinges installed at their ends; (ii)~two soft propulsors that propel the microswimmer forward; (iii)~two transmissions that amplify the output displacements generated by the robot's actuators into the two significantly larger stroke angles required to excite the two soft propulsors of the swimmer; (iv)~a convex hull made of CF that provides a buoyancy force to support the microswimmer on water; (v)~an \mbox{11-mAh} \mbox{$3.7$-V} \mbox{$507$-mg} \mbox{single-cell} \mbox{Li-Ion} battery (Powerstream~\mbox{GM$301014$H}); and, (vi)~the \mbox{custom-built} PCB described in~(a). \textbf{(d)}~Swimming mode of the VLEIBot\textsuperscript{++}\hspace{-0.4ex}. The two bare \mbox{SMA-based} microactuators undulate the tails of the swimmer to produce the hydrodynamic forces required for forward propulsion. \label{FIG02}}
\vspace{-2ex}
\end{figure}

\vspace{-1ex}
\section{The VLEIBot\textsuperscript{++}: An Autonomous Subgram Swimmer}
\label{SECTION02}
\vspace{-2ex}
\subsection{Design and Fabrication of the VLEIBot\textsuperscript{++}} 
\label{SECTION02.1}
\vspace{-1ex}
The design and procedure used to fabricate a \mbox{VLEIBot\textsuperscript{++}\hspace{-0.3ex}~prototype} are shown in \mbox{Fig.\,\ref{FIG02}}. The main element that enables tetherless \mbox{SMA-based} actuation for this swimmer is the \mbox{custom-built} PCB depicted in~\mbox{Figs.\,\ref{FIG02}(a)~and~(b)}. As seen in \mbox{Fig.\,\ref{FIG02}(a)}, the PCB is fabricated in three steps. In \mbox{Step\,1}, sheets of \textit{\mbox{copper-clad}~FR4}~(CuFR4) are laser etched to remove areas of the \mbox{Cu-coating} and thus create the patterns designed for each side of the PCB. In \mbox{Step\,2}, the two sides of the PCB are \mbox{pin-aligned} and adhered together with a sheet of Pyralux adhesive by applying pressure and heat inside a curing oven. In \mbox{Step\,3}, the \mbox{off-the-shelf} elements composing the interconnected power, actuation, and computation circuits are installed on the two sides of the PCB using conductive silver epoxy (\mbox{MG-Chemicals~$8331$D}) and cured inside an oven. Schematics of the two sides of the fabricated PCB are shown in~\mbox{Fig.\,\ref{FIG02}(b)}. The top side of the PCB includes a \textit{micro controller unit} (MCU)---Microchip Technology \mbox{PIC$16$F$18326$-I/JQ}---and an \textit{inertial measurement unit} (IMU)---STMicroelectronics \mbox{LSM$6$DSOXTR}---which respectively perform onboard computation and sensing. We included programming ports to the circuit design, which are required to upload the operational \mbox{C-code} that runs the MCU; the programming is implemented using an MPLabs \mbox{PicKit\,$4$} \mbox{in-circuit} debugger tool and the \mbox{MPLabs~X~IDE~v$6.20$} developer environment. The bottom side of the PCB includes two \textit{\mbox{metal-oxide-semiconductor} \mbox{field-effect} transistors}~(MOSFETs), which function as switches that open and close the \mbox{current-flow} pathways to the SMA actuators when triggered by \textit{\mbox{pulse-width} modulation} (PWM) signals generated by the MCU. As seen in \mbox{Fig.\,\ref{FIG02}(b)}, the bottom side of the PCB also includes a regulator that stabilizes the voltage delivered by the \mbox{Li-Ion} battery to power the MCU and IMU. The two sides of the PCB are connected to each other through cylindrical \textit{vias} filled with conductive silver epoxy. Also, decoupling capacitors and \mbox{pull-up/down} resistors were included in the PCB design to ensure proper circuit functionality. 

The assembly of a \mbox{VLEIBot\textsuperscript{++}\hspace{-0.3ex}~prototype} is graphically described in~\mbox{Fig.\,\ref{FIG02}(c)}. This robot consists of six main types of components: \mbox{(i)~two} bare \mbox{SMA-based} \mbox{microactuators} of the type described in~\cite{VLEIBot_2024} that drive the propulsors of the swimmer; \mbox{(ii)~two} soft tails that generate the hydrodynamic forces required for locomotion through \textit{fluid-structure interaction} (FSI); \mbox{(iii)~two} planar \mbox{four-bar} linkage transmissions that amplify the displacement outputs of the actuators into large stroke angles that undulate the two soft tails of the swimmer; \mbox{(iv)~a} convex hull made of CF that provides a container to store the system's battery and generates a buoyancy force that works in parallel with the surface tension of water to maintain the swimmer afloat; \mbox{(v)~an} \mbox{11-mAh} \mbox{$3.7$-V} \mbox{$507$-mg} \mbox{single-cell} \mbox{Li-Ion} battery (Powerstream~\mbox{GM$301014$H}) that enables the swimmer to operate continuously and autonomously for about $20\,\ts{min}$; and, \mbox{(vi)~the} \mbox{custom-built} PCB illustrated in \mbox{Figs.\,\ref{FIG02}(a)~and~(b)}. To complete the assembly procedure, all the components of the prototype are connected together and secured using \textit{cyanoacrylate} (CA) glue (\mbox{Loctite\,$414$}) and the swimmer's hull is sealed by lining the inside and outside of all the structure's seams with CA glue. The SMA wires that drive the actuators are made of nitinol with a nominal transition temperature of \mbox{$90\,^{\circ}\ts{C}$} and diameter of \mbox{$0.0381\,\ts{mm}$} (Dynalloy~$90\,^{\circ}\ts{C}$~HT). A schematic that explains the basic swimming mode of the \mbox{VLEIBot\textsuperscript{++}\hspace{-0.3ex}~is} shown in \mbox{Fig.\,\ref{FIG02}(d)}. As discussed in~\cite{VLEIBot_2024}, when the two soft tails of the propulsors are flapped, the resulting undulations produce through FSI the hydrodynamic forces that propel the swimmer forward. By independently modulating the two PWM signals that excite the two propulsors of the swimmer, we can vary the magnitude and direction of the total thrust propelling the system. As a result, \textit{two-dimensional} (2D) controllability is achieved, as empirically demonstrated through the \mbox{feedback-controlled} swimming tests performed using the \mbox{VLEIBot\textsuperscript{+}\hspace{-0.3ex}~presented} in~\cite{VLEIBot_2024}. 
\begin{figure}[t!]
\vspace{-0.2ex}  
\begin{center}    
\includegraphics[width=0.98\textwidth]{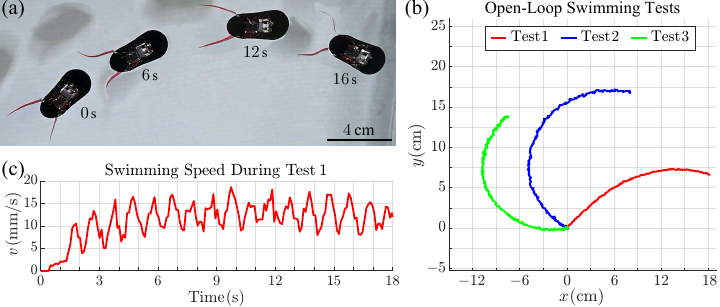}
\end{center}
\vspace{-5ex}    
\caption{\textbf{\mbox{Open-loop} swimming experiments of the VLEIBot\textsuperscript{++}\hspace{-0.4ex}.} \textbf{(a)} Photographic composite of frames taken at \mbox{$6$-s} intervals from video footage of \mbox{open-loop} swimming \mbox{Test\,$1$}. \mbox{\textbf{(b)}~$2$D~trajectories} of the VLEIBot\textsuperscript{++}\hspace{-0.4ex}, $\left\{x,y\right\}$, during \mbox{open-loop} swimming Tests\,$1$,~$2$,~and~$3$. In this plot, we placed the beginning of the trajectories at $\left\{0,0\right\}$ for the purpose of comparison. \mbox{\textbf{(c)}~Swimming} speed, $v$, during \mbox{Test\,$1$}. In this test, we measured average and maximum speeds of $11.9\,\ts{mm/s}$ and $18.7\,\ts{mm/s}$, respectively. Video footage of these swimming experiments can be seen in the accompanying supplementary movie. This movie is also available at \url{https://wsuamsl.com/resources/ISRR2024movie.mp4}. \label{FIG03}}
\vspace{-2ex}
\end{figure}

\subsection{Swimming Experiments of the VLEIBot\textsuperscript{++}} 
\label{SECTION02.2}
\vspace{-1ex}
To assess and demonstrate the functionality, basic controllability, and performance of a \mbox{$900$-mg} \mbox{VLEIBot\textsuperscript{++}\hspace{-0.3ex}~prototype} during autonomous operation, we conducted \mbox{feedforward-controlled} swimming tests. \mbox{Fig.\,\ref{FIG03}} presents experimental data from three tests in which the swimmer was programmed to excite its bare \mbox{SMA-based} microactuators using PWM signals with a frequency of $1\,\ts{Hz}$ and \textit{duty~cycle} (DC) of $5\,\%$. Here, \mbox{Fig.\,\ref{FIG03}(a)} shows a photographic composite of frames taken at intervals of $6\,\ts{s}$ from overhead footage of \mbox{Test\,$1$}. We estimated the trajectories of the tested prototype during swimming, through \mbox{frame-by-frame} video analyses, using a \mbox{custom-programmed} Matlab script. The trajectories of the swimmer during the three performed \mbox{open-loop} tests are shown in \mbox{Fig.\,\ref{FIG03}(b)}. Here, it can be observed that the swimmer repeatedly turned right in the three presented experiments, which indicates an inherent \mbox{right-hand} bias most likely produced by minor fabrication errors. From these simple tests, we conclude that for the \mbox{VLEIBot\textsuperscript{++}\hspace{-0.3ex}~to} stably follow trajectories in the 2D space, feedback control methods, such as those introduced in~\cite{VLEIBot_2024}, must be implemented. Additionally, we believe that turning biases can be eliminated by improving the propulsors' consistency of fabrication, a matter of current and further research in our laboratory. An important characteristic of any AUV---regardless of its size---is its agility, which is determined by the maximum achievable locomotion speed and turning rates. To estimate the VLEIBot\textsuperscript{++}'s forward speed, $v$, from the experimental data, we used a \mbox{discrete-time} derivative algorithm and a \mbox{one-second} \mbox{moving-average} digital filter. \mbox{Fig.\,\ref{FIG03}(c)} shows the swimmer's speed throughout \mbox{Test\,$1$}; in this case, we measured an average speed of \mbox{$11.9\,\ts{mm/s}$} (\mbox{$0.29\,\ts{Bl/s}$}) and a maximum speed of \mbox{$18.7\,\ts{mm/s}$} (\mbox{$0.46\,\ts{Bl/s}$}). The observed swimming functionality and measured performance of the tested prototype are very preliminary but highly promising. Overall, the \mbox{VLEIBot\textsuperscript{++}\hspace{-0.3ex}~represents} a significant step towards the creation of \mbox{insect-scale} AUVs in terms of propulsion, actuation, control, and computation. One major technical challenge that we still must overcome in order to develop a fully functional \mbox{insect-scale} AUV is efficient \mbox{low-power} underwater microactuation. We address this issue in the next section.

\vspace{-1ex}
\section{A New $13$-mg Low-Power Underwater Actuator} 
\label{SECTION03}
\vspace{-2ex}
Most actuators based on the thermal excitation of SMAs exhibit highly dissimilar behaviors in air and water due to the significantly different \mbox{heat-transfer} coefficients of these two media. Specifically, while the bare \mbox{SMA-based} microactuators that drive the \mbox{VLEIBot\textsuperscript{++}\hspace{-0.3ex}~can} function efficiently with the power electronics and \mbox{energy-storage} methods discussed in Section\,\ref{SECTION02} when operated in air, no \mbox{SMA-based} microactuator capable of functioning efficiently underwater has been developed to this date. In this section, we present experimental data regarding the power consumption of a typical bare \mbox{$10$-mg} \mbox{SMA-based} microactuator of the type used to drive \mbox{VLEIBot\textsuperscript{++}\hspace{-0.3ex}~prototypes}, state the identified challenges that need to be addressed to create \mbox{low-power} underwater \mbox{SMA-based} microactuation, and present design predictions empirically tested through the development of a new \mbox{$13$-\ts{mg}} \mbox{power-efficient} \mbox{SMA-based} microactuator for underwater operation conceived to drive \mbox{insect-scale} AUVs. 
\begin{figure}[t!]
\vspace{2ex}
\begin{center}
\includegraphics[width=0.98\textwidth]{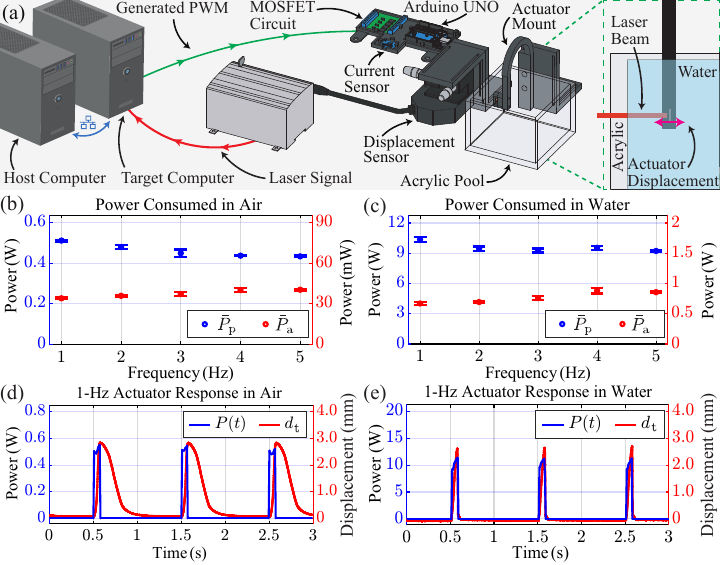}
\end{center}
\vspace{-5ex}
\caption{\textbf{\mbox{Power-consumption} characterization of bare \mbox{SMA-based} \mbox{actuator}.} \mbox{\textbf{(a)}~Experimental} setup used in the characterization experiments. A Mathworks Simulink \mbox{Real-Time} \mbox{host--target} system, equipped with a National Instruments \mbox{PCI-$6229$} \mbox{AD/DA} board, is used to generate the PWM signal that triggers the \mbox{MOSFET-based} circuit (\mbox{four-channel} \mbox{YYNMOS-$4$}) that excites the tested actuator; a current sensor (\mbox{Adafruit~INA$260$} in conjunction with an \mbox{Arduino~UNO}) measures and records the current that flows through the SMA wire that drives the actuator; a laser displacement sensor (Keyence~\mbox{LK-$031$}) measures the instantaneous actuation output and the corresponding data are recorded using the \mbox{host--target} system at a sampling rate of \mbox{$5$\,kHz}. We performed the \mbox{in-air} characterization first; then, we performed the underwater characterization using an acrylic pool filled with water. In~\textbf{(b)}~and~\textbf{(c)}, respectively corresponding to operation in air and water, each blue data point indicates the mean of five $P_{\ts{p}}$ values, $\bar{P}_{\ts{p}}$, and associated ESD corresponding to five \mbox{back-to-back} experiments, for the five different PWM pairs, \mbox{$\left\{f,\ts{DC}\right\}$}, considered here; and, each red data point indicates the mean of five $P_{\ts{a}}$ values, $\bar{P}_{\ts{a}}$, and associated ESD corresponding to the same five \mbox{back-to-back} experiments. Operating in air at \mbox{$f = 1$\,Hz} and \mbox{$\ts{DC} = 7\,\%$}, we measured average and peak power consumptions on the orders of \mbox{$40$\,mW} and \mbox{$0.5$\,W}, respectively. Operating underwater at \mbox{$f = 1$\,Hz} and \mbox{$\ts{DC} = 7\,\%$}, we measured average and peak power consumptions on the orders of \mbox{$800$\,mW} and \mbox{$10.6$\,W}, respectively. \mbox{\textbf{(d)--(e)}}~Show, in red, the responses of the actuator to a \mbox{$1$-Hz} \mbox{$7$-percent} PWM voltage while operating in air and water, respectively; and, in blue, the corresponding instantaneous power consumptions. Video footage of the tested bare \mbox{SMA-based} actuator operating in both air and water can be viewed in the accompanying supplementary movie. This movie is also available at \url{https://wsuamsl.com/resources/ISRR2024movie.mp4}. \label{FIG04}}
\vspace{-2ex}
\end{figure}

\subsection{Bare \mbox{SMA-Based} Microactuator During Underwater Operation} 
\label{SECTION03.1}
\vspace{-1ex}
To measure the power consumed by the tested \mbox{$10$-mg} bare \mbox{SMA-based} microactuator while operating in both air and water, we used the experimental setup illustrated in~\mbox{Fig.\,\ref{FIG04}(a)}. In this setup, a Mathworks Simulink \mbox{Real-Time} \mbox{host--target} system, equipped with a National Instruments \mbox{PCI-$6229$} \textit{\mbox{analog-digital/digital-analog}} (AD/DA) board, is employed to generate the PWM signal that triggers the \mbox{MOSFET-based} circuit (\mbox{four-channel} \mbox{YYNMOS-$4$}) that excites the tested actuator through tether wires, according to the same method \mbox{presented~in~\cite{VLEIBot_2024}}. During the tests, we utilize an intensity sensor (\mbox{Adafruit~INA$260$} in conjunction with an \mbox{Arduino~UNO}) and a laser displacement sensor (\mbox{Keyence~LK-$031$}) to measure the instantaneous current that flows through the SMA wire that drives the actuator and resulting instantaneous actuation output, respectively. With the purpose of measuring the device's power consumption accurately, we minimize the power dissipated through the tether wires by using five \mbox{$52$-AWG} conductors in parallel. This configuration provides an \mbox{energy-efficient} electrical connection that is sufficiently flexible not to affect actuation performance. Following the \textit{ceteris~paribus} principle, in order to physically ensure the same actuation response in air and water, we supply the tested device with an \mbox{\textit{on}-height} voltage---specifically, $2.7\,\ts{V}$ in air and $12\,\ts{V}$ in water---that nominally produces the same \mbox{low-frequency} displacement output (\mbox{$\sim\hspace{-0.4ex}3\,\ts{mm}$}) during a PWM cycle in both media. Through simple experiments, we determined that using a laser sensor to measure displacement through acrylic and water produces an attenuated measurement. To correct this issue, we empirically identified a static mapping between the \textit{perturbed} displacement measurement, $d_{\ts{p}}$---sensed through acrylic and water as seen in~\mbox{Fig.\,\ref{FIG04}(a)}---and the \textit{true} displacement, $d_{\ts{t}}$---sensed through air. For the particular setup used in the experiments presented here, we determined that \mbox{$d_{\ts{t}} = 1.58 \cdot d_{\ts{p}}$}. 

We computed the instantaneous electrical power, $P$, consumed by the tested bare \mbox{SMA-based} actuator using the measured current and resistance of the \mbox{NiTi} SMA wire (Dynalloy~$90\,^{\circ}\ts{C}$~HT, with a diameter of \mbox{$0.0381\,\ts{mm}$}), whose value we determined empirically together with that of the tether wires (\mbox{$\sim\hspace{-0.4ex}12\,{\rm\Omega}$}). As seen in \mbox{Figs.\,\ref{FIG04}(b)~and~(c)}, we computed the power consumption, in air and water, for five different PWM excitations; namely, those defined by the \mbox{frequency-DC} pairs, \mbox{$\left\{f,\ts{DC}\right\}$}, in the matching sets \mbox{$f \in \left\{1,2,3,4,5\right\}\,\hspace{-0.6ex}\ts{Hz}$} and \mbox{$\ts{DC} \in \left\{7,8,9,10,10 \right\}\,\hspace{-0.6ex}\%$}. For each test, we computed the peak power, $P_{\ts{p}}$, by averaging all the maximum values of $P$---which coincide with the times at which the exciting PWM signal is in the \mbox{\textit{on}-state}---over a \mbox{steady-state} period of $30\,\ts{s}$; similarly, we obtained the average power, $P_{\ts{a}}$, by computing the mean of $P$ over a \mbox{steady-state} period of $30\,\ts{s}$. In \mbox{Figs.\,\ref{FIG04}(b)~and~(c)}, each blue data point indicates the mean of five $P_{\ts{p}}$ values, $\bar{P}_{\ts{p}}$, and associated \textit{experimental standard deviation} (ESD) corresponding to five \mbox{back-to-back} experiments, for the five different PWM pairs, \mbox{$\left\{f,\ts{DC}\right\}$}, considered here; and, each red data point indicates the mean of five $P_{\ts{a}}$ values, $\bar{P}_{\ts{a}}$, and associated ESD corresponding to the same five \mbox{back-to-back} experiments. For \mbox{in-air} operation, $\bar{P}_{\ts{p}}$ is on the order of $0.5\,\ts{W}$ at a PWM frequency of $1\,\ts{Hz}$ and slightly decreases as the frequency increases; $\bar{P}_{\ts{a}}$ is on the order of $40\,\ts{mW}$ for frequencies above $3$\,Hz and slightly less at lower frequencies. For underwater operation, $\bar{P}_{\ts{p}}$ is on the order of \mbox{$10\,\ts{W}$} at a PWM frequency of $1$\,Hz and slightly decreases as the frequency increases; $\bar{P}_{\ts{a}}$ is on the order of $800\,\ts{mW}$ for frequencies greater than $3\,\ts{Hz}$ and slightly less at lower frequencies. This increase of about $1900\,\%$ in average power consumption clearly indicates that the bare \mbox{SMA-based} microactuators developed to drive \mbox{VLEIBot\textsuperscript{++}\hspace{-0.3ex}~prototypes} do not represent a feasible possibility for \mbox{insect-scale} AUVs because large amounts of energy would need to be carried onboard to sustain reasonably long mission times. These findings prompted us to address the problem of underwater power consumption in \mbox{SMA-based} microactuation. We propose and discuss a solution next. 

\subsection{A New \mbox{Low-Power} Underwater SMA-Based Microactuator} 
\label{SECTION03.2}
\vspace{-1ex}
From simple observation of the actuator responses presented in~\mbox{Figs.\,\ref{FIG04}(d)}~and~(e), we can conclude that during underwater operation, the tested bare \mbox{SMA-based} actuator rapidly returns to its rest position when the exciting PWM signal transitions from the \mbox{\textit{on}-state} to the \mbox{\textit{off}-state}. Assuming that the stress experienced by the SMA wire that drives a microactuator is similar in air and water, the noticeably faster actuation responses to excitation variations during underwater operation compared to those occurred in air---observed in~\mbox{Fig.\,\ref{FIG04}(e)}---are explained by the substantially more rapid cooling of the SMA material in water than in air. This phenomenon occurs because the \mbox{heat-transfer} coefficient of water is significantly larger than that of air at normal conditions of temperature and pressure (\mbox{$20\,^{\circ}\ts{C}$}; \mbox{$1\,\ts{atm}$}). This observation suggests that \mbox{power-efficient} underwater \mbox{SMA-based} microactuation can be achieved by reducing the \mbox{heat-transfer} coefficient of the local media surrounding the SMA wire driving the system to a level similar to that of air at $20\,^{\circ}\ts{C}$. Here, we use fundamental \mbox{heat-transfer} analyses to assess the feasibility of a design with the potential to achieve \mbox{low-power} actuation in water, which we discuss next.

\subsubsection{\mbox{Heat-Transfer} Analysis.} We use a lumped model to describe the \mbox{heat-transfer} dynamics of a generic piece of material~\cite{HeatTransferBook}. Namely,
\begin{align}
\frac{dT}{dt}= \frac{1}{m C_{\ts{p}}} \left( Q_{\ts{in}} - Q_{\ts{out}} \right),
\label{EQN01}   
\end{align} 
in which $T$ is the temperature of the material; $m$ is the mass of the piece; $C_{\ts{p}}$ is the specific heat capacity of the material; $Q_{\ts{in}}$ is the rate at which heat is transferred into the piece; and, $Q_{\ts{out}}$ is the rate at which heat is transferred out of the piece, estimated as
\begin{align}
Q_{\ts{out}} = \frac{T_{\ts{h}} - T_{\ts{c}}}{R_{\ts{eq}}},
\label{EQN02}
\end{align}
in which $T_{\ts{h}}$ and $T_{\ts{c}}$ are respectively the hotter and colder temperatures of the two points at the edges of the equivalent thermal resistance $R_{\ts{eq}}$; and, \mbox{$R_{\ts{eq}} = R_{\ts{conv}} + R_{\ts{cond}}$}, where $R_{\ts{conv}}$ and $R_{\ts{cond}}$ are lumped convection and conduction thermal resistances, respectively. Using the notion of thermal resistance network for cylindrical tubes~\cite{HeatTransferBook}, in the case considered here, 
\begin{align}
R_{\ts{cond}} = \frac{\ln\left(r_{\ts{in}}/r_{\ts{out}}\right)}{2\pi Lk} \quad \ts{and} \quad R_{\ts{conv}} = \frac{1}{\bar{h}A_{\ts{c}}},
\label{EQN03}
\end{align} 
where $r_{\ts{in}}$ is the inner radius of the modeled tube; $r_{\ts{out}}$ is the outer radius of the tube; $L$ is the length of the tube; $k$ is the thermal conductivity of the material; $\bar{h}$ is the estimated \mbox{heat-transfer} coefficient between the material and surrounding environment; and, $A_{\ts{c}}$ is the contact area through which heat is transferred. 
\begin{figure}[t!]
\vspace{-0.2ex}
\begin{center}
\includegraphics[width=0.98\textwidth]{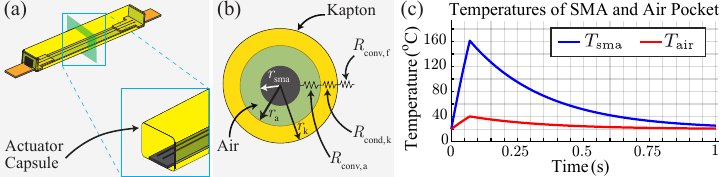}
\end{center}
\vspace{-4ex}
\caption{\textbf{Proposed solution for underwater actuation.} \textbf{(a)}~Design of encapsulated \mbox{SMA-based} microactuator capable of underwater operation. \textbf{(b)}~\mbox{One-dimensional} \mbox{heat-transfer} model of an SMA wire surrounded by an air pocket contained by a \mbox{Kapton-made} capsule. $R_{\ts{cond},\ts{k}}$ is the thermal resistance for conduction through the Kapton membrane, $R_{\ts{conv},\ts{a}}$ and $R_{\ts{conv},\ts{f}}$ are the thermal resistances for convection through the air and external fluid (air or water), respectively. \textbf{(c)}~\mbox{Numerical-simulation} results for the temperatures over time of both the driving SMA wire, $T_{\ts{sma}}$, and air cavity, $T_{\ts{air}}$, during a \mbox{$1$-Hz} \mbox{$7$-percent} \mbox{PWM-excited} actuation cycle of the encapsulated microactuator during underwater operation. \label{FIG05}}
\vspace{-2ex}
\end{figure}

The value of $\bar{h}$ depends on both the properties of the surrounding fluid and kinematics of the corresponding flow. For gases, typical values are in the range of \mbox{$2$~to~$250\,\ts{W}/(\ts{m}^2\cdot\ts{K})$}; for liquids, typical values are in the range of $50$~to~\mbox{$20000\,\ts{W}/(\ts{m}^2\cdot\ts{K})$}~\cite{HeatTransferBook}. This significant difference between the \mbox{heat-transfer} coefficients in air and water is the reason behind the considerable increase in power consumption when the tested bare \mbox{SMA-based} actuator is operated underwater. To address this problem, we present the \mbox{design-based} solution depicted in \mbox{Fig.\,\ref{FIG05}}. This design is based on the use of an insulating capsule made of a thin layer of Kapton (\mbox{$7.5$\,{\textmu}m}) that contains air, as depicted in~\mbox{Fig.\,\ref{FIG05}(a)}. The insulating pocket of air significantly reduces the local \mbox{heat-transfer} coefficient of the media surrounding the SMA wire. The membrane of Kapton serves as an impermeable barrier that prevents leaking and also offers low resistance to heat transfer, which keeps the air in the capsule at a temperature close to that of the surrounding water. To assess its feasibility, we use the idealization shown in~\mbox{Fig.\,\ref{FIG05}(b)}. As seen, we assume an encapsulated SMA wire with an annular air pocket enclosed by a Kapton membrane, in which \mbox{$r_{\ts{sma}} = 0.01905$\,mm} is the radius of the modeled bare SMA wire; \mbox{$r_{\ts{a}} - r_{\ts{sma}}=0.1$\,mm} is the annular thickness of the air pocket; \mbox{$r_{\ts{k}}-\left(r_{\ts{a}}+r_{\ts{sma}}\right)=0.0127$\,mm} is the thickness of the Kapton membrane; $R_{\ts{conv},\ts{a}}$ is the thermal resistance across the air pocket; $R_{\ts{cond},\ts{k}}$ is the thermal resistance across the Kapton membrane; and, $R_{\ts{conv},\ts{f}}$ is the thermal resistance from the Kapton membrane's surface to the surrounding fluid. 

The temperature dynamics of both the SMA wire and annular air pocket can be described using the model specified by~(\ref{EQN01}). In this case, the average rate at which heat is transferred into the SMA material through Joule heating during a \mbox{PWM-excited} actuation cycle is \mbox{$Q_{\ts{in},\ts{sma}} = 10^{-2} \cdot \ts{DC} \cdot R_{\ts{sma}} \cdot I^2$}, where \mbox{$R_{\ts{sma}} = 8.9\,{\rm \Omega}$} and \mbox{$I=125\,\ts{mA}$} are the electrical resistance of the SMA wire and current flowing through it during the \mbox{\textit{on}-section} of the exciting PWM cycle, respectively. In this analysis, we assume a PWM signal with a frequency of $1\,\ts{Hz}$ and DC of $7\,\%$. Next, to estimate $R_{\ts{conv},\ts{a}}$,
we use the convection equation in (\ref{EQN03}) with a \mbox{heat-transfer} coefficient of \mbox{$210\,\ts{W}/(\ts{m}^2\cdot\ts{K})$} and wire length of \mbox{$10$\,mm}. To estimate $R_{\ts{conv},\ts{f}}$, we assume the same length of \mbox{$10$\,mm} and a \mbox{heat-transfer} coefficient in water of \mbox{$5000\,\ts{W}/(\ts{m}^2\cdot\ts{K})$}. To estimate $R_{\ts{cond},\ts{k}}$, we use the conduction equation in (\ref{EQN03}), assuming a wire length of \mbox{$10$\,mm} and a thermal conductivity of \mbox{$0.2\,\ts{W}/(\ts{m}\cdot\ts{K}$)}. According to the modeling approach depicted in~\mbox{Fig.\,\ref{FIG05}(b)}, heat is first transferred from the SMA material into the air pocket; then, heat is transferred through the Kapton membrane into the surrounding water. \mbox{Fig.\,\ref{FIG05}(c)} shows the simulated temperature evolution of both the SMA material and air pocket over one PWM cycle, assuming that the actuator is immersed in water with a temperature of $20\,^{\circ}\ts{C}$. These simulations were implemented in Matlab using a discretized version of (\ref{EQN01}) with a step size of \mbox{$10^{-8}$\,s}. To simulate the temperature of the SMA material, we used a mass of~\mbox{$7.35\cdot10^{-8}\,\ts{kg}$} and specific heat capacity of \mbox{$836.8\,\ts{J}/(\ts{kg}\cdot\ts{K})$}; to simulate the temperature of the annular air pocket, we used a mass of \mbox{$5.31\cdot10^{-10}\,\ts{kg}$} and specific heat capacity of \mbox{$1005\,\ts{J}/(\ts{kg}\cdot\ts{K})$}. Here, it can be seen that the piece of \mbox{NiTi} SMA material (Dynalloy~$90\,^{\circ}\ts{C}$~HT, with a diameter of $0.0381\,\ts{mm}$) surpasses its nominal transition temperature while the temperature of the air cavity remains relatively close to that of the surrounding water ($20\,^{\circ}\ts{C}$). These results indicate that by using a sealed capsule filled with air to provide a local medium for the SMA wires driving the microactuator, we can achieve levels of power consumption similar to those measured during the characterization experiments of the tested bare \mbox{SMA-based} microactuator operating in air, while achieving similar mechanical functionality. 
\begin{figure}[t!]
\vspace{-0.2ex}
\begin{center}    
\includegraphics[width=0.98\textwidth]{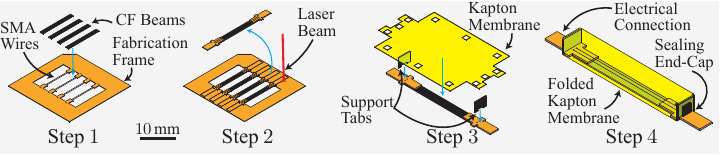}
\end{center}
\vspace{-4ex}
\caption{\textbf{Fabrication of \mbox{SMA-based} encapsulated actuator.} In \mbox{Step\,$1$}, \mbox{NiTi} SMA wires (Dynalloy\,\mbox{$90\,^{\circ}\ts{C}$~HT}; \mbox{$0.038$\,mm}) and CF beams are installed on a precut fabrication frame. In \mbox{Step\,$2$}, a \mbox{355-nm} \mbox{$3$-W} laser (Photonics~Industries~\mbox{DCH-$355$-$3$}) is used to release the assembled actuators with integrated \mbox{electrical-connection} tabs. In \mbox{Step\,$3$}, for each bare actuator fabricated in \mbox{Step\,$2$}, CF support tabs are attached to the device's structure, and a foldable sheet made of Kapton is perforated and patterned using the \mbox{high-power} laser. In \mbox{Step\,$4$}, for each actuator, the precut Kapton membrane is folded over the support tabs and the CF beam; to seal the ends of the resulting capsule, CF \mbox{end-caps} are installed and fixed using CA glue. The seam of the Kapton capsule is folded over the CF beam and sealed using flexible silicone adhesive (\mbox{Smooth-On}~\mbox{Sil-Poxy}). \label{FIG06}}
\vspace{-2ex}
\end{figure}

\subsubsection{Design and Fabrication of Microactuator for Underwater Operation.}
The fabrication process of the proposed \mbox{low-power} encapsulated \mbox{SMA-based} microactuator is depicted in \mbox{Fig.\,\ref{FIG06}}. As seen, actuators are fabricated in sets of four and the procedure consists of four steps. In \mbox{Step\,$1$}, the SMA wires (Dynalloy\,\mbox{$90\,^{\circ}\ts{C}$~HT}; \mbox{$0.038$\,mm}) and CF beams composing the actuators are installed on a fabrication frame according to the process described in~\cite{SMALLBug_2020}. \mbox{In Step\,$2$}, the four actuators with their electrical connections are cut from the fabrication frame using a \mbox{high-power} laser (Photonics~Industries~\mbox{DHC-$355$-$3$}). In \mbox{Step\,$3$}, CF support tabs are attached to the bare actuators fabricated in \mbox{Step\,$2$}, and foldable sheets made of Kapton are perforated and patterned using the \mbox{high-power} laser. In \mbox{Step\,$4$}, for each actuator, a precut Kapton membrane is folded over the support tabs and the CF beam; to seal the ends of the resulting capsule, CF \mbox{end-caps} are installed and fixed using CA glue. After the Kapton membrane is folded around the CF beam, the seam of the capsule is sealed using flexible silicone adhesive (\mbox{Smooth-On}~\mbox{Sil-Poxy}) to prevent leaking during actuation. \mbox{Cu-FR4} tabs cut from the fabrication frame in \mbox{Step\,$2$} allow electrical connection to the SMA wires through the Kapton membrane. After testing the functionality of the actuator in air, the electrical connections are sealed using a coating of CA glue (\mbox{Loctite~$414$}) before underwater testing. After fabrication, the proposed encapsulated \mbox{SMA-based} microactuator weighs only $13\,\ts{mg}$, a $30\,\%$ increase from the original bare \mbox{SMA-based} actuator; we tested its functionality and performance in both air and water, and measured its power consumption in both media through a series of experiments that we discuss next.
\begin{figure}[t!]
\vspace{-0.2ex}
\begin{center}
\includegraphics[width=0.98\textwidth]{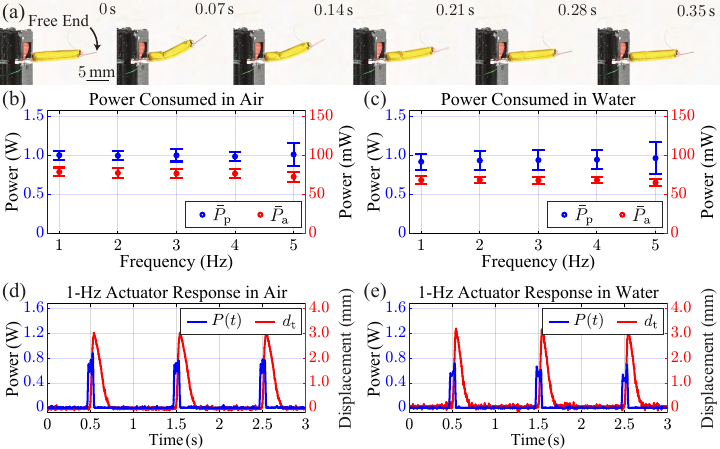}
\end{center}
\vspace{-5ex}   
\caption{\textbf{\mbox{Power-consumption} characterization of encapsulated \mbox{SMA-based} \mbox{actuator}.} During characterization, the tested device was excited using the PWM signals defined by the pairs \mbox{$\left\{f,\ts{DC}\right\}$} in the matching sets \mbox{$f \in \left\{1,2,3,4,5\right\}\,\hspace{-0.6ex}\ts{Hz}$} and \mbox{$\ts{DC} \in \left\{7,8,9,10,10 \right\}\,\hspace{-0.6ex}\%$}, with an \mbox{\textit{on}-height} of \mbox{$3.3$\,V}. \textbf{(a)}~Photographic sequence, taken from lateral video footage, of one actuation cycle of the tested encapsulated microactuator operating underwater at \mbox{$1$\,Hz}. In~\textbf{(b)}~and~\textbf{(c)}, respectively corresponding to operation in air and water, each blue data point indicates the mean of five $P_{\ts{p}}$ values, $\bar{P}_{\ts{p}}$, and associated ESD corresponding to five \mbox{back-to-back} experiments, for the five different PWM pairs, \mbox{$\left\{f,\ts{DC}\right\}$}, considered here; and, each red data point indicates the mean of five $P_{\ts{a}}$ values, $\bar{P}_{\ts{a}}$, and associated ESD corresponding to the same five \mbox{back-to-back} experiments. Operating in air at \mbox{$f = 1$\,Hz} and \mbox{$\ts{DC} = 7\,\%$}, we measured average and peak power consumptions on the orders of \mbox{$80$\,mW} and \mbox{$1$\,W}, respectively. Operating underwater at \mbox{$f = 1$\,Hz} and \mbox{$\ts{DC} = 7\,\%$}, we measured average and peak power consumptions on the orders of \mbox{$70$\,mW} and \mbox{$0.97$\,W}, respectively. \mbox{\textbf{(d)--(e)}}~Show, in red, responses of the actuator to a \mbox{$1$-Hz} \mbox{$7$-percent} PWM voltage while operating in air and water, respectively; and, in blue, the corresponding instantaneous power consumptions. Video footage of the proposed encapsulated \mbox{SMA-based} actuator operating in both air and water can be viewed in the accompanying supplementary movie. This movie is also available at \url{https://wsuamsl.com/resources/ISRR2024movie.mp4}.
\label{FIG07}}
\vspace{-2ex}
\end{figure}

\subsubsection{Power Characterization of Proposed \mbox{$\bs{13}$-mg} Encapsulated Actuator.} To characterize the power consumption of the proposed \mbox{$13$-mg} encapsulated \mbox{SMA-based} actuator in both air and water, we used the experimental setup already described in \mbox{Section\,\ref{SECTION03.1}}; however, we replaced the \mbox{Adafruit~INA$260$} current sensor with a \mbox{Pololu~$05$AU} measuring device, whose outputted voltage was read directly using the \mbox{AD/DA} board of the setup. As in the power characterization of the bare \mbox{SMA-based} actuator presented in Section\,\ref{SECTION03.1}, we electrically connected the tested encapsulated actuator to its two ends using five \mbox{$52$-AWG} wires in parallel. Through empirical observations, we determined that an \mbox{\textit{on}-height} excitation voltage of \mbox{$3.3$\,V} induces the tested encapsulated actuator to generate output displacements comparable to those generated by the tested bare \mbox{SMA-based} actuator (see \mbox{Section\,\ref{SECTION03.1}}). The experimental data obtained from testing an encapsulated prototype are summarized in~\mbox{Fig.\,\ref{FIG07}}. Here, \mbox{Fig.\,\ref{FIG07}(a)} shows a photographic sequence, taken from lateral video footage, of one actuation cycle of the tested device operating underwater and excited by a PWM signal with a frequency of \mbox{$1$\,Hz}, DC of $7\,\%$, and \mbox{\textit{on}-height} voltage of \mbox{$3.3$\,V}. In~\mbox{Figs.\,\ref{FIG07}(b)~and~(c)}, respectively corresponding to operation in air and water, each blue data point indicates the mean of five $P_{\ts{p}}$ values, $\bar{P}_{\ts{p}}$, and associated ESD corresponding to five \mbox{back-to-back} experiments, for the five different PWM pairs, \mbox{$\left\{f,\ts{DC}\right\}$}, considered here; and, each red data point indicates the mean of five $P_{\ts{a}}$ values, $\bar{P}_{\ts{a}}$, and associated ESD corresponding to the same five \mbox{back-to-back} experiments. \mbox{Figs.\,\ref{FIG07}(d)~and~(e)} present, in red, the actuator responses, $d_{\ts{t}}$, excited by a PWM signal with a frequency of \mbox{$1$\,Hz}, DC of $7\,\%$, and \mbox{\textit{on}-height} voltage of \mbox{$3.3$\,V}, during operation in air and water, respectively; estimates of the corresponding instantaneous consumed electrical powers are shown in blue. As seen, we measured average and peak power consumptions at \mbox{$1$\,Hz} on the orders of \mbox{$80$\,mW} and \mbox{$1$\,W}, respectively. As expected by design, we measured a similar power consumption when operating underwater; namely, average and peak power consumptions at \mbox{$1$\,Hz} on the orders of \mbox{$70$\,mW} and \mbox{$0.97$\,W}, respectively. While the \mbox{in-air} average power consumption at \mbox{$1$\,Hz}, relative to that measured for the tested bare actuator, increases from about \mbox{$40$\,mW} to about \mbox{$80$\,mW}---an increase of approximately \mbox{$100\,\%$}---the underwater average power consumption decreases from about \mbox{$800$\,mW} to about \mbox{$70$\,mW}---a drastic decrease of approximately \mbox{$91\,\%$}. These results constitute the main contribution of this paper. 

\vspace{-1ex}
\section{Conclusions and Future Directions}
\label{SECTION04}
\vspace{-2ex}
We presented a new \mbox{$900$-mg} surface swimmer, the VLEIBot\textsuperscript{++}\hspace{-0.4ex}, which employs two \mbox{low-power} bare \mbox{SMA-based} microactuators---in conjunction with a lightweight \mbox{custom-built} PCB that provides sensing, computation, and electronic capabilities---to locomote autonomously. A single charge of an onboard \mbox{Li-Ion} battery sustains about $20\,\ts{min}$ of untethered operation. Through swimming tests, we determined that the \mbox{VLEIBot\textsuperscript{++}\hspace{-0.3ex}~can} locomote in open loop at speeds of up to \mbox{$18.7\,\ts{mm}/\ts{s}$}~($0.46\,\ts{Bl/s}$) while operating its actuators at $1\,\ts{Hz}$. We presented a study of the \mbox{power-consumption} requirements of a bare \mbox{SMA-based} microactuator of the type used to drive the \mbox{VLEIBot\textsuperscript{++}\hspace{-0.3ex}~while} operating in both air and water; we determined that the average power required to function underwater increases by about $1900\,\%$ with respect to that required for \mbox{in-air} operation. These results prompted us to develop a new \mbox{$13$-mg} \mbox{low-power} HWD \mbox{SMA-based} actuator for underwater operation that consumes an average power on the order of $80\,\ts{mW}$ when operating in air or water. This figure represents a decrease of about $91\,\%$ in underwater power consumption with respect to that of the tested bare \mbox{SMA-based} microactuator. As part of our research program, we will integrate the new underwater \mbox{SMA-based} actuation technology presented here into swimmers of the \mbox{VLEIBot\textsuperscript{++}\hspace{-0.3ex}~type} to create the very first \mbox{insect-scale} AUVs.

\vspace{4ex}
\hspace{-4ex} \textbf{Acknowledgments.} 
This work was partially funded by the Washington State University (WSU) Foundation and the Palouse Club through a Cougar Cage Award to \mbox{N.\,O.\,P.-A.}, the US National Science Foundation (NSF) through~\mbox{Award\,$2244082$}, and the Morgan Family Charitable Trust through a direct gift to \mbox{N.\,O.\,P.-A.}; additional funding was provided by the WSU Voiland College of Engineering and Architecture through a start-up fund to \mbox{N.\,O.\,P.-A}.

\bibliographystyle{IEEEtran}
\bibliography{references}
\end{document}